\documentclass[journal,twoside,web]{ieeecolor}
\usepackage{generic}
\usepackage{cite}
\usepackage{amsmath,amssymb,amsfonts}
\usepackage{algorithmic}
\usepackage{graphicx}
\usepackage{algorithm,algorithmic}
\usepackage{hyperref}
\hypersetup{hidelinks=true}
\usepackage{textcomp}
\usepackage{bm}
\usepackage{epstopdf}
\usepackage{graphicx}
\usepackage{diagbox}
\usepackage{booktabs}
\usepackage{multirow}
\usepackage{siunitx}
\usepackage{cite}
\usepackage[justification=centering]{caption}
\def\BibTeX{{\rm B\kern-.05em{\sc i\kern-.025em b}\kern-.08em
    T\kern-.1667em\lower.7ex\hbox{E}\kern-.125emX}}
\begin{document}
\title{A Field Calibration Approach for Triaxial MEMS Gyroscopes Based on Gravity and Rotation Consistency}
\author{Yaqi Li, Li Wang, Zhitao Wang, Xiangqing Li, Steven Weidong Su 
    \thanks{Yaqi Li, Zhitao Wang, Xiangqing Li and Steven Weidong Su are with the College of Medical information and Artificial intelligence Shandong First Medical University Shandong Academy of Medical Science 6699 Qingdao Road, Jinan, 250024, Shandong, China}
    \thanks{Li Wang is with the CSIRO Space \& Astronomy, PO Box 1130, Bentley, WA 6102, Australia}
    \thanks{}
    \thanks{\textsl{(corresponding author:Steven Weidong Su)}}}

\maketitle

\begin{abstract}
This paper developed an efficient method for calibrating triaxial MEMS gyroscopes, which can be effectively utilized in the field environment. The core strategy is to utilize the criterion that the dot product of the measured gravity and the rotation speed in a fixed frame remains constant. To eliminate the impact of external acceleration, the calibration process involves separate procedures for measuring local gravity and rotation speed. Moreover, unlike existing approaches for auto calibration of triaxial sensors that often result in nonlinear optimization problems, the proposed method simplifies the estimation of the gyroscope scale factor by employing a linear least squares algorithm. Extensive numerical simulations have been conducted to analyze the proposed method's performance in calibrating the six-parameter triaxial gyroscope model, taking into consideration measurements corrupted by simulated noise. Experimental validation was also carried out using two commercially available MEMS inertial measurement units (LSM9DS1) and a servo motor. The experimental results effectively demonstrate the efficacy of the proposed calibration approach.
\end{abstract}

\begin{IEEEkeywords}
    Tri-axial gyroscope, in-field calibration.
\end{IEEEkeywords}

\section{Introduction}
\label{sec:introduction}
\IEEEPARstart{M}{icro}-electro system(MEMS), comprised of accelerometer, gyroscope, and magnetometers, find wide application in health monitoring equipment\cite{9090976}\cite{bo2022imu} \cite{majumder2020wearable}, smartphones\cite{khedr2017smartphone}\cite{sun2022indoor}, personal consumer device\cite{han2016extended}\cite{gemelli2020low} and more, owing to their affordability, lower power consumption, and compact dimensions\cite{zhanshe2015research}\cite{al2023imu}. Nevertheless, when compared to the accelerometer and expensive gyroscope(such as fiber gyroscope \cite{lefevre2022fiber}\cite{wang2022recent}), MEMS gyroscopes demonstrate low accuracy. Many factors contribute to the MEMS gyroscope errors and the errors of the gyroscope can be categorized into two types: systematic errors and stochastic errors. Systematic errors refer to those with a fixed value, such as zero offset, axis misalignment, and scale factors\cite{zhang2019error}\cite{ghanipoor2020toward}. However, the stochastic errors including flicker error and rate random walk are not systematic in nature\cite{mi9080373}\cite{zhang2020impact}. Hence, calibrating the stochastic errors by straightforward method poses significant challenges \cite{FAZELINIA2024115136}. Instead of eliminating stochastic errors, it is more feasible to enhance the gyroscope accuracy by regularly calibration to eliminate the systemic errors. Among various systematic errors, in-run bias stability and scale factor stand out as the predominant sources of error during the working process. To address the offset, a method that maintains gyroscope stability for a duration and averages static data can be employed. Calibrating zero offset involves subtracting the average from all data points. The scale factor is pivotal in gyroscope calibration, representing the primary factor influencing accuracy enhancement. So devising a swift and efficient method to calibrate the MEMS gyroscope scale factor stands as a pressing and formidable endeavor.

The matter of gyroscope calibration has collected significant attention\cite{9295651}\cite{harindranath2023systematic}. The ordinary triaxial gyroscope calibration approach\cite{chatfield1997fundamentals} can achieve high accuracy but the complex calibration procedure and the need for costly equipment render it unsuitable for applications beyond the laboratory setting. A pivotal and frequently expensive step in these methods is acquiring comprehensive information about the true motions applied on the IMU. The Calibration methods which not need expensive equipment are presented at \cite{wu2017gyroscope}  \cite{wang2021efficient} \cite{6413274} \cite{9739788}. One of these approaches \cite{wu2017gyroscope} introduces a Calibration method that utilizes a magnetometer. This technique requires a consistent magnetic field for gyroscope calibration which make the method is not suitable for the certain in-field application. Also, this method calls for a high-precision robotic arm, a requirement that limits its suitability beyond controlled laboratory environment. About\cite{wang2021efficient} and \cite{zhou2020novel}, the method involved comparing several different gyroscope orientation and obtaining the error through integration. However, this method can only calibrate the overall scale factor and is unable to calibrate specific angle velocity point. The method given in \cite{6413274} calibrates the gyroscope by camera, which can provide the images and orientation information. The initial requirement of this method is aligning the image frame with the IMU frame, while the second requirement is the utilization of high performance computing equipment due to the inclusion of image data. So it is not suitable for the edge computing devices. In \cite{9739788}, temporal convolutional network is employed to calibration the MEMS IMU gyroscope, while it can compensate for multiple gyroscope errors simultaneously, the neural network requires substantial computing power. Additionally, due to the unexplainable nature of deep neural networks, this method cannot guarantee stability in all cases.

The calibration method proposed in this paper achieves the goal of rapidly and efficiently calibrating MEMS gyroscope in field environment. On the one hand, this method does not require high-precision equipment, needing only a cheap and commonly available servo motor, as shown in Figure \ref{equipment}. On the other hand, the method proposed in this paper does not face installation alignment or secondary installation issues. Therefore, this method is not affected by installation errors\cite{looney2015basics}\cite{lu2022new} during the calibration process. We summarize the contributions of this paper are as follows: Firstly, we introduce a rapid and efficient method capable of calibrating the MEMS gyroscope scale factor. Secondly, the proposed method does not necessitate expensive calibration equipment; only a cheap servo motor capable of controlling rotational speed is required. Thirdly, our method eliminates the need for secondary gyroscope installation and avoids the calibration errors associated with traditional calibration processes involving multiple installations.

\section{Calibration Method}

In the following section, we present the proposed calibration approach, along with its underlying mathematical framework.

The most commonly employed gyroscope models in calibration processes typically include two types: the 9-parameter and 6-parameter models. A critical component to be calibrated in the 9-parameter model is the scale factor matrix, which is represented as follows: 
\[
\mathbf{K}=
\begin{bmatrix}
	K_{xx} & K_{xy} & K_{xz} \\
	K_{yx} & K_{yy} & K_{yz} \\
	K_{zx} & K_{zy} & K_{zz}
\end{bmatrix}.
\]
However, it is noteworthy that the influence of the off-diagonal elements 
\[
K_{xy}, K_{xz}, K_{yx}, K_{yz}, K_{zx}, K_{zy}
\]
on the calibration outcomes can often be considered negligible. Consequently, we focus on the 6-parameter model in this study, as it emerges as the preferable option for MEMS (Micro-Electro-Mechanical Systems) gyroscope sensors. To ensure the accuracy of our exposition, additional parameters will be introduced as necessary. In the section pertaining to gyroscope calibration, we define the uncalibrated angular velocity as \(\mathbf{G^{m} \in \mathbb{R}^{3}}\), and the calibrated value, \(\mathbf{G^{r} \in \mathbb{R}^{3}}\), is given by:
\[
\mathbf{G^{r}} = \mathbf{K} \mathbf{G^{m}} + \mathbf{b}
\]
where $\mathbf{K}$ is the scale factor matrix of the 6-parameter model. In order to simplify our discussion, we simply use $K_x, K_y, K_z$ stands for $K_{xx}, K_{yy}, K_{zz}$ respectively, i.e., 
\[
\mathbf{K} = \begin{bmatrix}K_{x} & 0 & 0 \\0 & K_{y} & 0 \\0 & 0 & K_{z} \\\end{bmatrix}
\]
and 
\[
\mathbf{b} = \begin{bmatrix} b_{x} & b_{y} & b_{z} \end{bmatrix}^{T}
\]
represents the bias. 

The acceleration vector is represented by 
\[
\mathbf{A} = \begin{bmatrix} A_{x} & A_{y} & A_{z} \end{bmatrix}^T.
\]
Given that the dot product between the acceleration vector and the calibrated gyroscope vector is constant, this relationship can be expressed as:
\begin{equation}\label{eq_0}
L = \mathbf{A} \cdot \mathbf{G^{r}} = 
\begin{bmatrix}
	A_{x} & A_{y} & A_{z}
\end{bmatrix}
\begin{bmatrix}
	G_{x}^{r} \\ G_{y}^{r} \\ G_{z}^{r}
\end{bmatrix} = A_{x}G_{x}^{r} + A_{y}G_{y}^{r} + A_{z}G_{z}^{r}
\end{equation}

To ensure the precision and fidelity of the calibration values, the measured acceleration vector, \(\mathbf{A^{m}} = \begin{bmatrix} A^{m}_{x} & A^{m}_{y} & A^{m}_{z} \end{bmatrix}^{T}\), is initially subjected to the calibration procedures outlined in our previous research \cite{ye2017efficient}. The resulting calibrated values are denoted as \(\mathbf{A^{c}} = \begin{bmatrix} A^{c}_{x} & A^{c}_{y} & A^{c}_{z} \end{bmatrix}^{T}\).

Based on Equation (\ref{eq_0}) and the calibrated acceleration vector \(\mathbf{A^{c}} = \begin{bmatrix} A^{c}_{x} & A^{c}_{y} & A^{c}_{z} \end{bmatrix}^{T}\), the constant \(L\) is defined as:

\begin{equation} \label{eq_1}
   	L = A^c_{x}(K_{x}G^{m}_{x} + b_{x}) + A^c_{y}(K_{y}G^{m}_{y} + b_{y}) + A^c_{z}(K_{z}G^{m}_{z} + b_{z})
\end{equation}

Equation (\ref{eq_1}) can be reformulated into a matrix representation, as shown in Equation (\ref{eq_2}), when there are 
$n$ sets of measurement data:

\begin{equation} \label{eq_2}
   	\begin{bmatrix}
   		A_{x1}^{c}G_{x}^{m} & A_{x1}^{c} & A_{y1}^{c}G_{y}^{m} & A_{y1}^{c} & A_{z1}^{c}G_{z}^{m} & A_{z1}^{c} \\
   		\cdots              & \cdots     & \cdots              & \cdots     & \cdots              & \cdots     \\
   		\vdots              & \vdots     & \vdots              & \vdots     & \vdots              & \vdots     \\
   		A_{xn}^{c}G_{x}^{m} & A_{xn}^{c} & A_{yn}^{c}G_{y}^{m} & A_{yn}^{c} & A_{zn}^{c}G_{z}^{m} & A_{zn}^{c}
   	\end{bmatrix}
   	\begin{bmatrix} K_{x}\\ b_{x}\\ K_{y}\\ b_{y}\\ K_{z}\\ b_{z} \end{bmatrix}  = \begin{bmatrix}
   		L \\ \vdots \\ L
   	\end{bmatrix}
\end{equation}

The bias \(\mathbf{b} = \begin{bmatrix} b_{x} & b_{y} & b_{z} \end{bmatrix}^{T}\) can be easily calibrated when the gyroscope is stationary. Thus, the primary considered parameter to be calibrated in this study is the scale factor matrix $\mathbf{K}=diag\{[K_{x}\, K_{y}\, K_{z}]\}$.
%
    
Based on Equation (\ref{eq_2}), the relationship can be simplified as:
\begin{equation}
    \bm{L = X\beta + \epsilon}
\end{equation}
where,
$$\mathbf{X} = \begin{bmatrix}
    A_{x1}^{c}G_{x}^{m} &  & A_{y1}^{c}G_{y}^{m} &  & A_{z1}^{c}G_{z}^{m} & \\
    \cdots              &      & \cdots     &      & \cdots              &      \\
    \vdots  &     & \vdots              &      & \vdots              &      \\
    A_{xn}^{c}G_{x}^{m} & & A_{yn}^{c}G_{y}^{m} & & A_{zn}^{c}G_{z}^{m} &  \\
\end{bmatrix}$$ and 
\[
\bm{\beta} = \begin{bmatrix}
{K_{x}} \\
{K_{y}} \\
{K_{z}}
\end{bmatrix}^{T}.
\]
The \(\epsilon\) represents the measurement noise, and here we assume it is white noise with zero mean. A least squares estimation algorithm can be formulated to calculate the parameter \(\bm \beta\) in Equation (\ref{eq_6}):
\begin{equation} \label{eq_6}
   	\bm{\beta = (X^{T}X)^{-1}X^{T}L}
\end{equation}
where the constant \(\bm{L}\) is related to the servo motor's rotating speed. More precisely, let
\begin{equation} \label{eq_7}
   	\bm{\hat{\beta}} = \begin{bmatrix} \hat{K_{x}} \\ \hat{K_{y}} \\ \hat{K_{z}} \end{bmatrix} = \bm{(X^{T} X)^{-1} X^{T}} \bm{1}
\end{equation}
where \(\bm{1}_{n \times 1}\) is a unit vector.

Then, the final scale factor
\[
\bm{\beta = \hat{\beta}}L = \begin{bmatrix} \hat{K_{x}} L \\ \hat{K_{y}} L \\ \hat{K_{z}} L \end{bmatrix}
\]
can be adjusted based on the servo motor's rotating speed. The equation is as shown below:
\begin{equation}
   	(G_{x}^{m}\hat{K}_{x} L)^{2} + (G_{y}^{m}\hat{K}_{y} L)^{2} + (G_{z}^{m}\hat{K}_{z} L)^{2} = \omega_n^2
\end{equation}
where the constant \(\omega_n\) represents the servo motor's rotating speed. Once \(L\) is calculated, the calibrated scale factor is
\[
\bm{K} = \begin{bmatrix} \hat{K}_{x} L \\ \hat{K}_{y} L \\ \hat{K}_{z} L \end{bmatrix}
\]

\section{Simulation validation}

The simulation process will be systematically outlined as follows, with calibration outcomes presented subsequently. Additionally, the impact of accelerometer errors and noise originating from IMU electronics on the calibration results will be investigated.

In the simulation, coordinate transformations will be employed to replicate the IMU's installation orientation and its associated rotation axis. Quaternions will be used to simulate the variations in the accelerometer components as the IMU rotates about its installation axis.

\subsection{Simulation Without Acceleration Error}
Initially, simulations will be conducted under ideal conditions, assuming a perfect accelerometer free of errors and influenced only by Gaussian noise. Based on these assumptions, the results will be presented as follows:
\begin{enumerate}
	\item The gyroscope scale factor is assumed to follow a uniform distribution \( U(0.9, 1.1) \), while the bias follows \( U(-2^{\circ}/s, 2^{\circ}/s) \). Typically, low-cost MEMS gyroscopes exhibit scale factors and biases within approximately \( \pm 10\% \) and \( \pm 2^{\circ}/s \), respectively.
	\item The accelerometer measurement noise is modeled as white noise, \( \mathcal{N}(0, 0.005^2) \).
	\item The variance of the rotation noise is proportional to 5\% of the current speed, following a Gaussian distribution \( \mathcal{N}(0, (0.05\omega)^2) \), and each sample of the noise is uncorrelated with the others. This simulates motor vibrations during operation.
	\item The proposed method in this paper requires only a single installation step for the entire calibration process, eliminating the need to account for installation errors. Any tilt angle introduced during installation is treated as random and does not require further adjustments.
\end{enumerate}

\begin{figure}
    \centering
    \includegraphics[width = 0.4\textwidth]{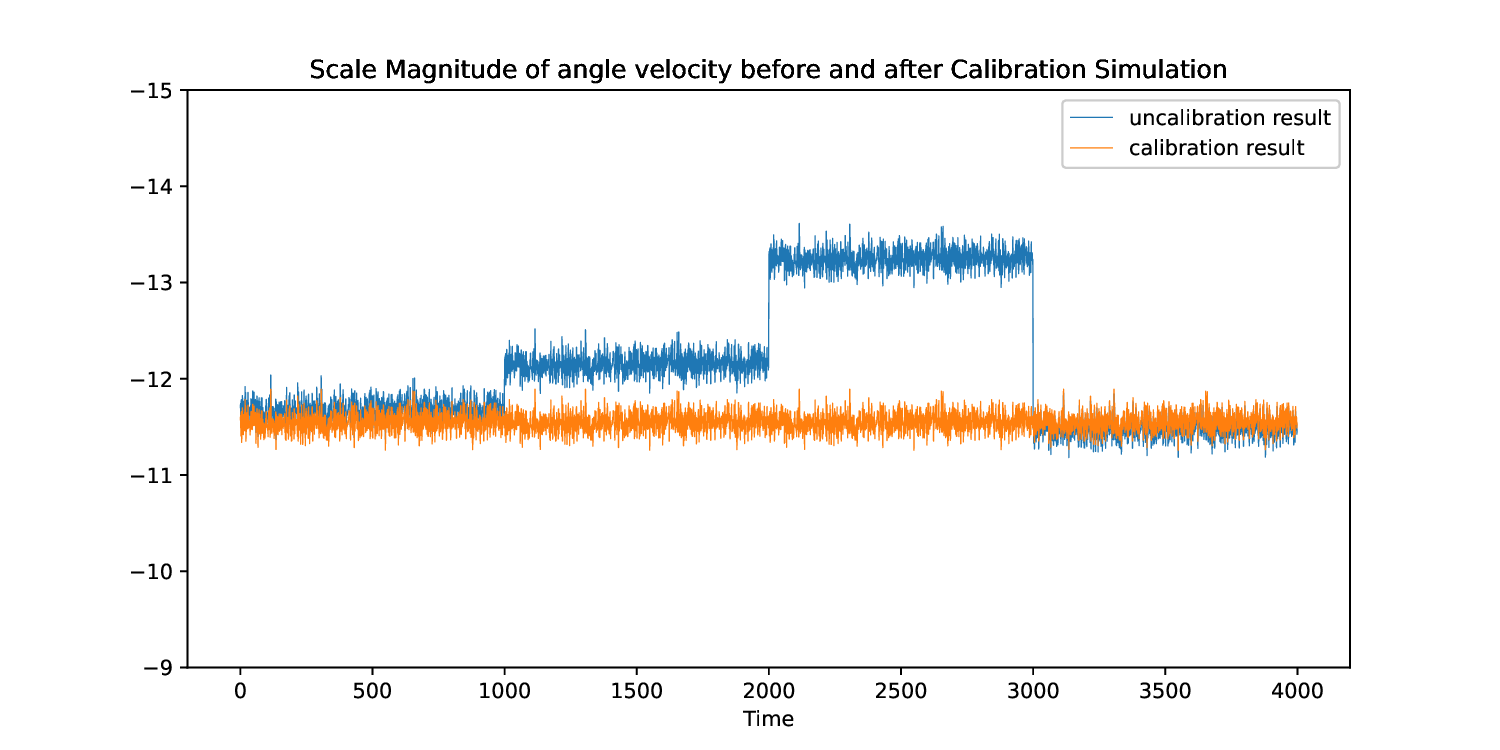}
    \caption{Magnitude of angular velocity before and after calibration in the simulation setting.}
    \label{simulationsimpledata}
\end{figure}

\begin{figure}
    \centering
    \includegraphics[width = 0.45\textwidth]{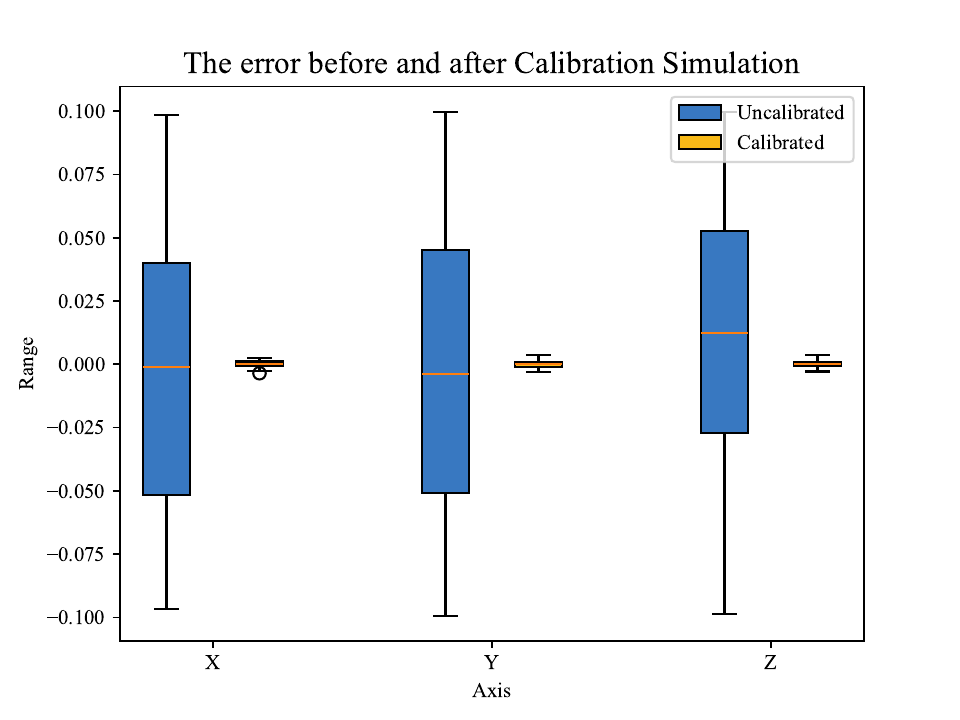}
    \caption{Error before and after calibration in the simulation.}
    \label{simulationwithouterror}
\end{figure}

\begin{table}
	\centering
	\caption{Mean, variance, and range of errors before and after calibrated}
	\begin{tabular}{ccccc}
		\toprule
		Status & Axis & Mean & Variance & Range\\ 
		\midrule 
		\multirow{3}*{Uncalibrated} & X & 0.0024 & 0.0035 & 0.9148\\
		~ & Y & 0.0039 & 0.0033 & 0.1943\\
		~ & Z & 0.0064 & 0.0030 & 0.1975\\
		\multirow{3}*{Calibrated} & X & 0.0001& $1.8815 \times 10^{-6}$ & 0.0067\\
		~ & Y & 0.0001& $2.3400 \times 10^{-6}$ & 0.0078\\
		~ & Z & 0.0001& $1.4951 \times 10^{-6}$ & 0.0063\\
		\bottomrule
	\end{tabular}
	\label{the table for Fig2}
\end{table}

\begin{figure}
	\centering
	\includegraphics[width = 0.5\textwidth]{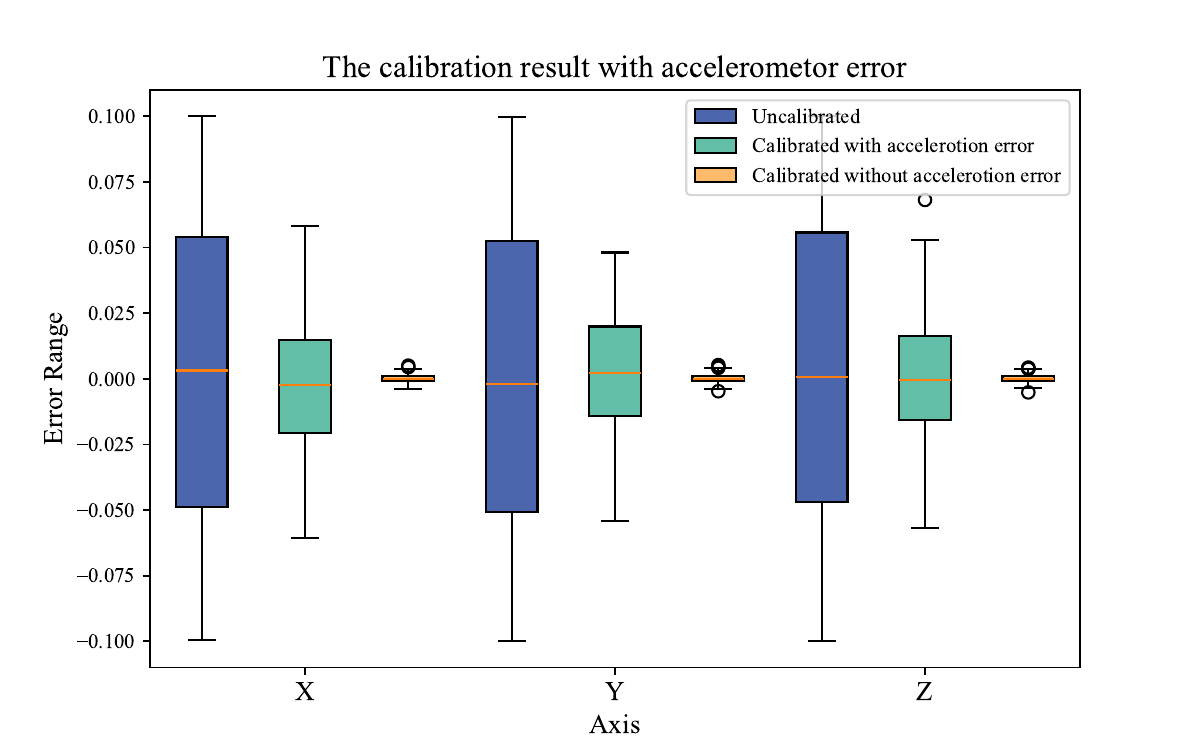}
	\caption{Comparison of Calibration result with and without accelerometer error}
	\label{boxplotaccerror}
\end{figure}

%

Based on the assumptions above, we generated scale factors and biases to simulate the actual gyroscope. Figure \ref{simulationsimpledata} illustrates the outcomes of the calibration process. It can be observed in Figure \ref{simulationsimpledata} that, after calibration, the dot product between the accelerometer vector and gyroscope vector becomes consistent.

To further validate the accuracy and stability of the method, we generated 100 sets of scale factors for the simulation. In each simulation, the scale factors and Gaussian noise were randomly generated, and the accelerometer's rotation angle was also random, while the axis of rotation remained stable. Box plots were used to analyze the error before and after calibration. The results are shown in Figure \ref{simulationwithouterror} and Table \ref{the table for Fig2}. The difference in error before and after calibration is significant. As seen in Figure \ref{simulationwithouterror} and Table \ref{the table for Fig2}, this method consistently calibrates the gyroscope's scale factor, even when the scale factor drifts randomly and the accelerometer's rotation angle is variable. This demonstrates that the proposed method exhibits a degree of robustness.

\subsection{Simulation with Acceleration Error}
In this subsection, simulation tests are conducted considering acceleration errors due to measurement inaccuracies in the accelerometers.
The assumptions for the gyroscope-related parameters remain the same as mentioned previously. The acceleration scale factor error follows a uniform distribution \( U(0.95, 1.05) \), while the biases follow a uniform distribution \( U(-0.005\, \text{m/s}^{2}, 0.005\, \text{m/s}^{2}) \), and the noise in the accelerometer data follows a normal distribution.

\begin{table}
	\caption{Calibration results with and without error}
	\centering
	\begin{tabular}{cccc}
		\toprule
		Axis & Uncalibrated error & CR with acc error & CR without acc error \\
		\midrule
		X    & -0.0165             & -0.0049           & -0.0027          \\
		Y    & 0.04406             & -0.0103           & -0.0015          \\
		Z    & -0.0999             & -0.0108           & -0.0009          \\
		\bottomrule
	\end{tabular}
	\label{thespecificsimualtionresult}
\end{table}

The simulation results are presented in Figure \ref{boxplotaccerror} and Table \ref{thespecificsimualtionresult}. A box plot is used to illustrate the distribution of the scale factor error in Figure \ref{boxplotaccerror}. When the gyroscope is not calibrated, the error distribution of its scale factor aligns with the assumed error distribution. After calibration using the accelerometer data with errors, the error range is reduced but not entirely eliminated, with a residual error of approximately 0.02 in the scale factor. However, when the accelerometer data is pre-calibrated using the method proposed by Lin Ye \cite{ye2017efficient}, and subsequently the gyroscope is calibrated using the method proposed in this study, a more accurate scale factor is obtained. The detailed data for a specific calibration process is shown in Table \ref{thespecificsimualtionresult}. In summary, pre-calibration of the accelerometer data is crucial for the effectiveness of this calibration method.

\subsection{The noise analysis}
In MEMS, noise is inevitable. If not properly managed, noise can significantly impact the performance of MEMS devices. There are two primary sources of noise in MEMS: external noise and internal noise, generated by external mechanical vibration and internal electronic components\cite{Mohd-Yasin_2010}. It will have a great impact on the stability of the system. In this subsection, we will analyze how noise affects the calibration results. 

\begin{figure*}
	\centering
	\includegraphics[width=1\textwidth]{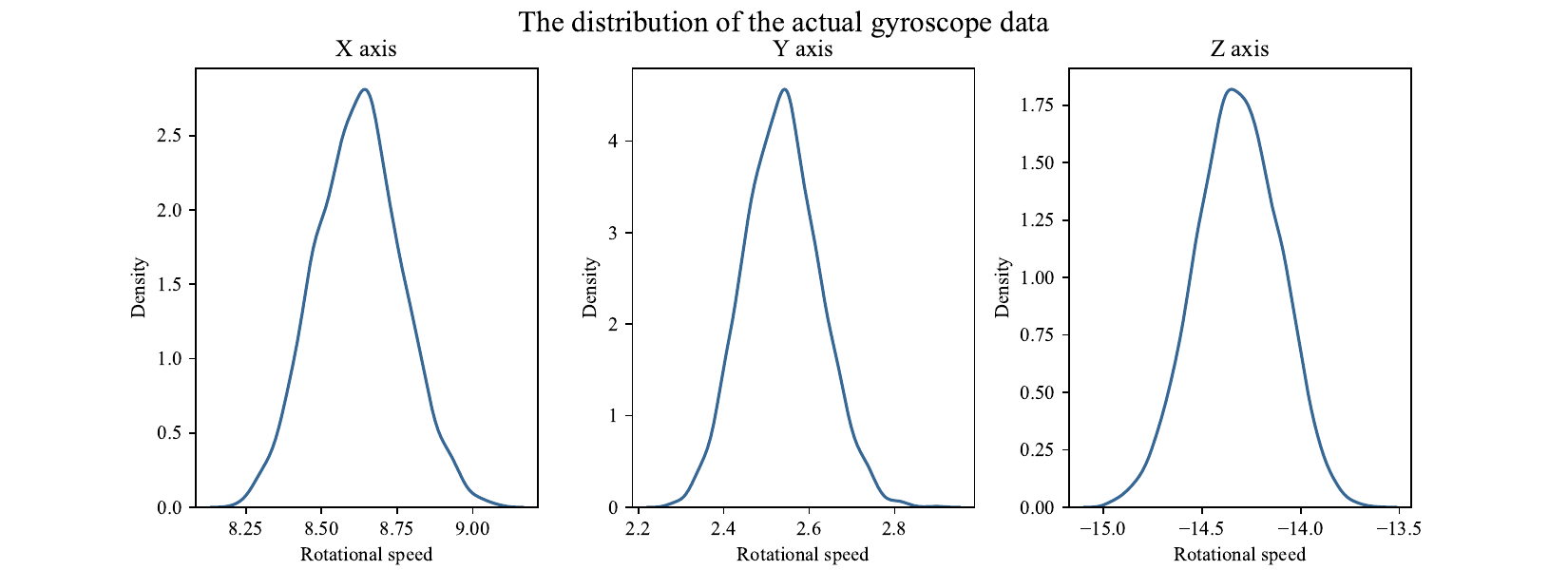}
	\caption{The distribution of the actual gyroscope data}
	\label{the kdeplot of actual gyro}
\end{figure*}

Initially, to ascertain whether the empirical data from the gyroscope adhere to a presumed normal distribution, a comprehensive analysis was conducted. The results of this analysis are delineated in Figure \ref{the kdeplot of actual gyro} and Table. \ref{distribution of actual gyro}. From the above chart, it can be concluded that the noise generated by the gyroscope under uniform rotation follows a normal distribution. 

\begin{table}
	\caption{The distribution information of actual gyroscope data}
	\centering
	\begin{tabular}{ccccc}
		\toprule
		 Axis & Mean & Variance & Kurtosis & Skewness \\
		\midrule
		 X & 10.002 & 0.027 & -0.188 & 0.084  \\
		 Y & 2.942 & 0.011 & -0.016 & 0.160    \\
		 Z & -16.613 & 0.060 & -0.213 &-0.069   \\
		\bottomrule
	\end{tabular}
	\label{distribution of actual gyro}
\end{table}

In the simulation, the drift scale factor of the gyroscope is assumed to be 
$[1.1, 0.9, 1.2]$. Regarding bias, the gyroscope is calibrated for a period of time before calibration to remove zero drift. The noise of gyroscope is assumed to be $\mathcal{N}(0, 0.01^2)$, $\mathcal{N}(0, 0.1^2)$, $\mathcal{N}[0,1^2]$, $\mathcal{N}[0, 10^2]$ respectively. And set the rotational speed to $10^{\circ}/s$.

\begin{figure}
	\centering
	\includegraphics[width = 0.55\textwidth, height=0.25\textheight]{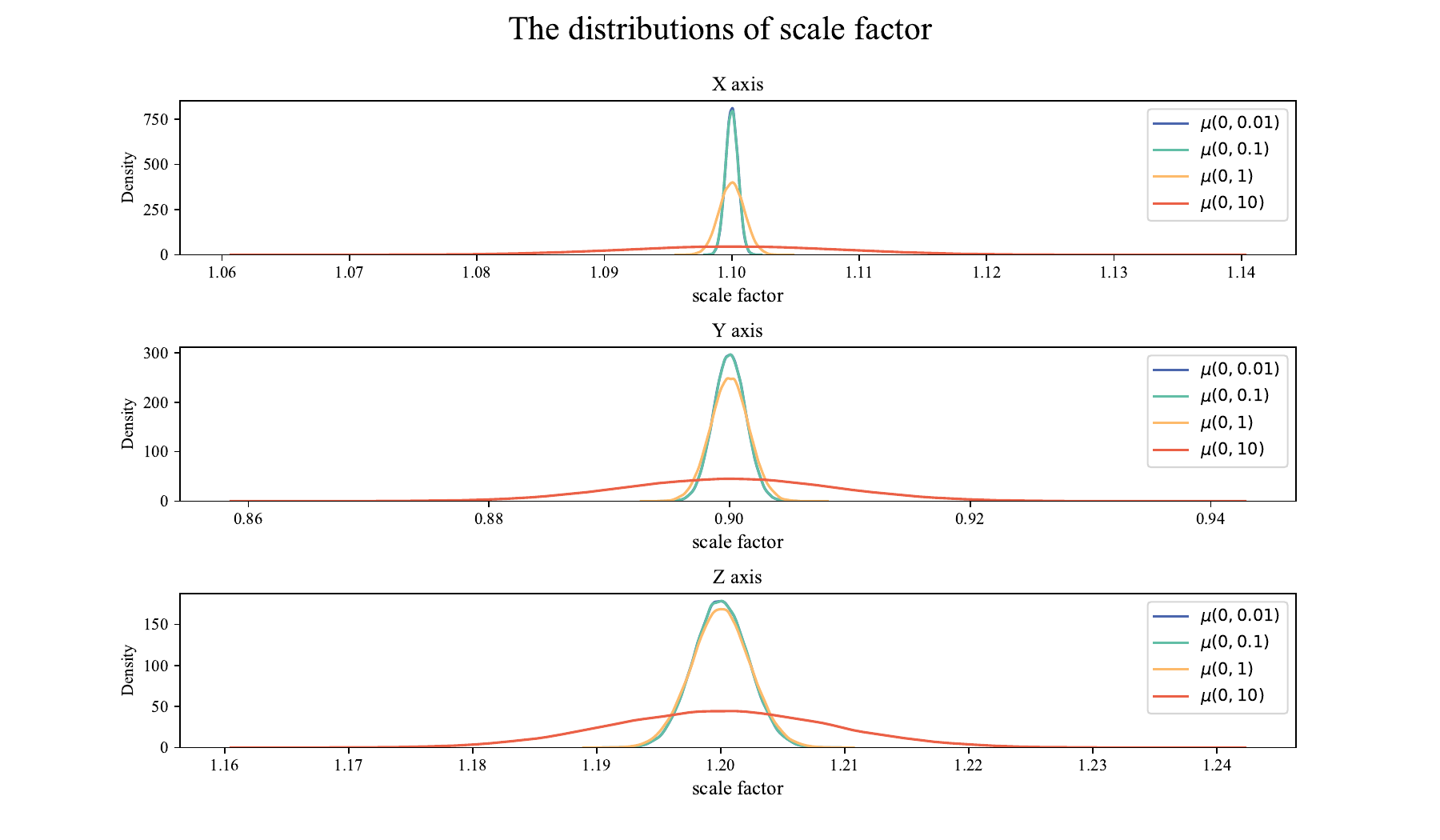}
	\caption{Distribution of scale factors under different noise influences}
	\label{scalefactordistribution}
\end{figure}

We use Python to simulate 100000 times for each noise condition. The distribution of scale factors for each axis is shown in Figure \ref{scalefactordistribution} and Table \ref{theinformationofscalefactor}. From the aforementioned figure and table, we can conclude that the variance of the Gaussian noise generated by the MEMS during operation has no significant effect on the distribution of the calibration results. Table \ref{theinformationofscalefactor} demonstrates that as the variance of the Gaussian noise increases gradually form 0.01 to 10, the kurtosis and skewness of the calibration result distribution remain essentially unchanged. Thus, the influence of varying Gaussian noise levels on the calibration result distribution can be considered negligible.

\begin{table}
	\caption{Mean, variance, kurtosis, and skewness of the scale factors generated by the simulation }
	\centering
	\begin{tabular}{cccccc}
		\toprule
		Axis & Variance of noise &Mean & Variance & Kurtosis & Skewness \\
		\midrule
		\multirow{4}*{X} & $\mathcal{N}(0,0.01^{2})$ &1.100 & \num{2.374e-7} & -0.006 & 0.007 \\
		~ & $\mathcal{N}(0,0.1^{2})$ & 1.100 & \num{2.447e-7} & -0.020 & 0.011 \\
		~ & $\mathcal{N}(0,1^{2})$ & 1.100 & \num{9.853e-7} & -0.005 & 0.004 \\
		~ & $\mathcal{N}(0,10^{2})$ & 1.099 & \num{7.519e-5} & 0.018 & -0.004 \\
		\multirow{4}*{Y} & $\mathcal{N}(0,0.01^{2})$ &0.899 & \num{1.758e-6} & -0.020 & 0.008 \\
		~ & $\mathcal{N}(0,0.1^{2})$ & 0.899 & \num{1.765e-6} & -0.020 & 0.007 \\
		~ & $\mathcal{N}(0,1^{2})$ & 0.900 & \num{2.509e-6} & -0.011 & -0.006 \\
		~ & $\mathcal{N}(0,10^{2})$ & 0.900 & \num{7.738e-5} & 0.002 & 0.007 \\
		\multirow{4}*{Z} & $\mathcal{N}(0,0.01^{2})$ &1.200 & \num{4.846e-6} & -0.019 & 0.019 \\
		~ & $\mathcal{N}(0,0.1^{2})$ &1.200 & \num{4.854e-6} & -0.019 & 0.019 \\
		~ & $\mathcal{N}(0,1^{2})$ & 1.200 & \num{5.562e-6} & 0.0002 & 0.004 \\
		~ & $\mathcal{N}(0,10^{2})$ & 1.199 & \num{7.980e-5} & 0.010 &-0.002 \\
		\bottomrule
	\end{tabular}
	\label{theinformationofscalefactor}
\end{table}

\subsection{Real-time experiments}
\begin{figure}
	\centering
	\includegraphics[width=3in]{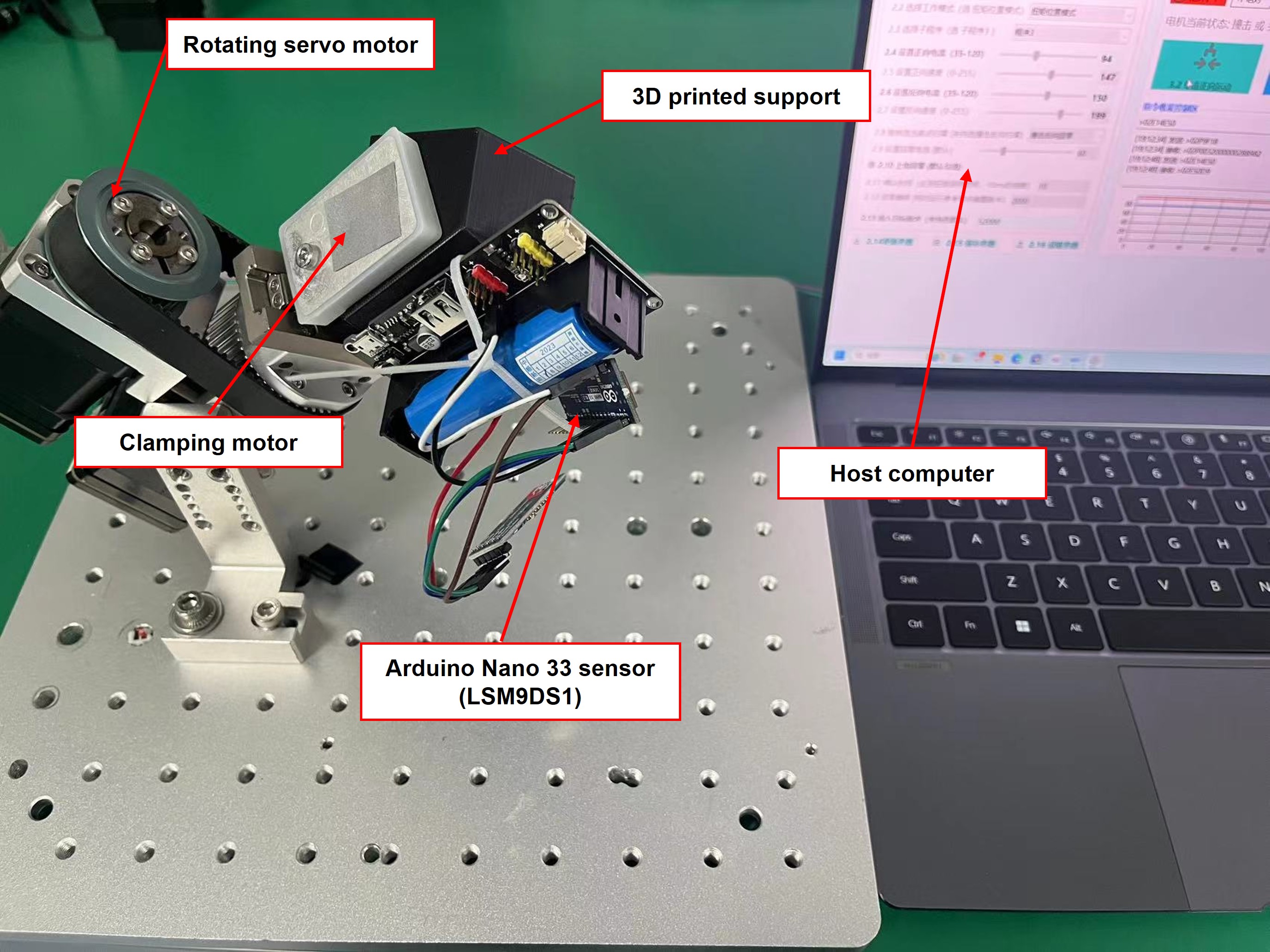}
	\caption{Experimental equipment consisting of two servo motors and a clamping device}
	\label{equipment}
\end{figure}

The experiment was conducted using a commercially available IMU device(LSM9DS1 from Arduino nano 33 BLE). We applied the proposed method to a experimental equipment consisting of two servo motor and a clamping device. The experimental device is shown in Figure \ref{equipment}. The digital signals were acquired and processes using an Arduino nano 33 BLE. Notably, the processed system did not require the use of high-precision turntables or other calibration equipment. During the experiment, the IMU sensor was rotated at a constant speed. While the IMU sensor was in motion, gyroscope data were collected. While maintaining a consistent axis of rotation, randomly reorient the IMU to obtain static acceleration from a minimum of four distinct positions. The complete process has a duration of approximately several minutes. The micro-controller recorded the data, which was subsequently transmitted to the computer through a serial port.

\begin{figure}
    \centering
    \includegraphics[width=0.5\textwidth,height= 0.22 \textheight]{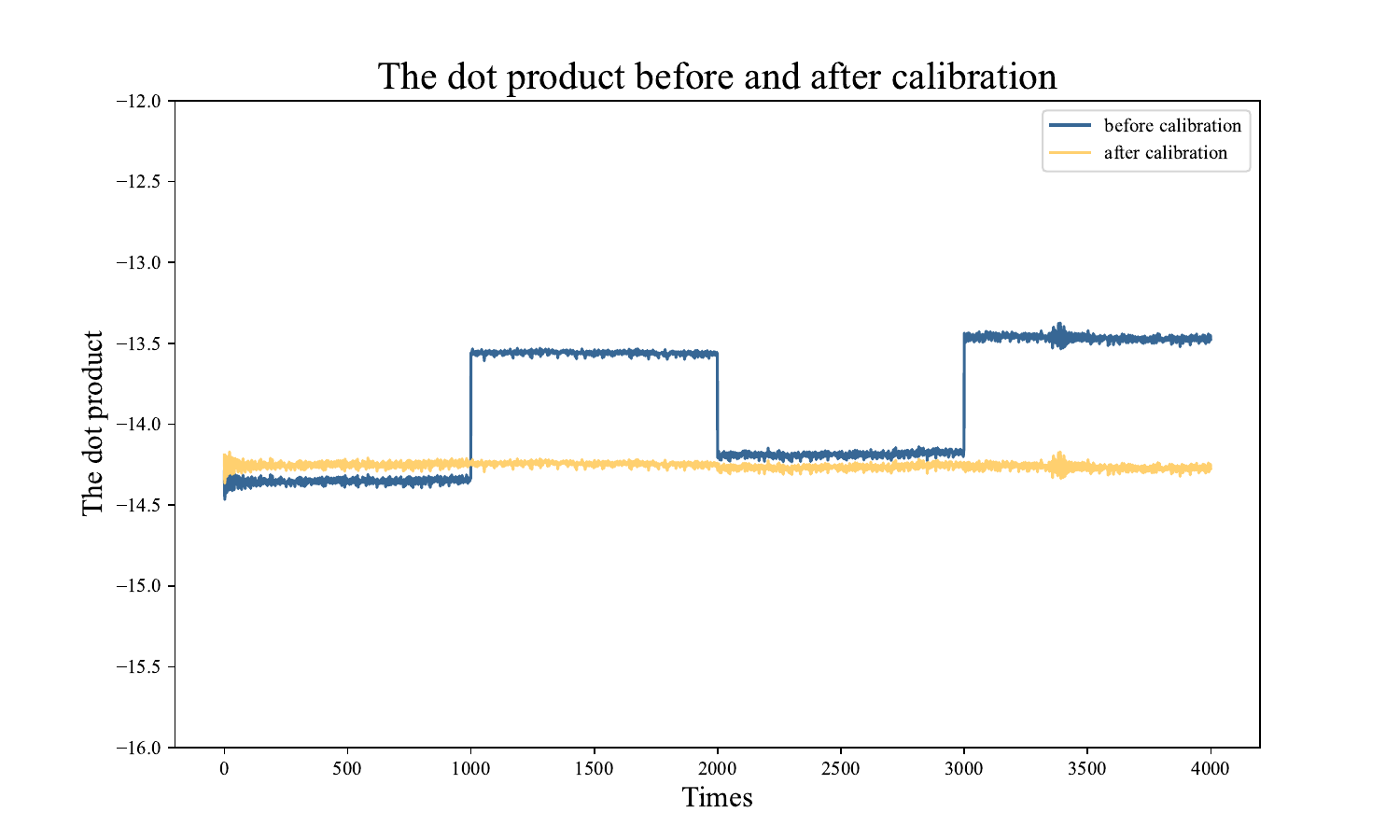}
    \caption{The dot product of before calibration and after calibration}
    \label{originaldata}
\end{figure}

\begin{figure}
    \centering
    \includegraphics[width=0.5 \textwidth, height=0.26 \textheight]{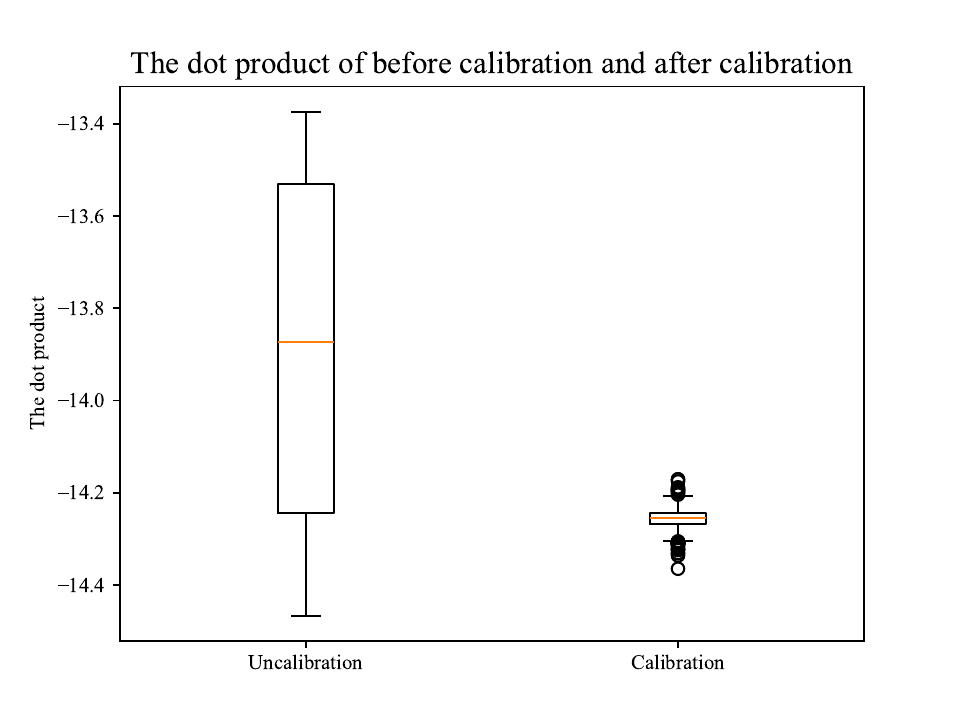}
    \caption{The boxplot depicts the dot product between the accelerometer vector and gyroscope vector including both uncalibration and calibration data}
    \label{theboxplotrange}
\end{figure}

\subsection{Calibration results}
In our experimental analysis, a specific rotational speed was selected for calibration using the proposed method. Specifically, a rotational speed of $20^{\circ}/s$ was chosen for the calibration experiment. The calibration results are shown in Figure \ref{originaldata} and \ref{theboxplotrange}. It is evident from the dot product between the accelerometer data and raw gyroscope data that the three axes exhibit varying degrees of drift. However, after calibration using the proposed method, the dot product becomes consistent across all axes.

\subsection{Analysis when the accelerometer is not stationary}
In this subsection, we will verify the assumption that the centrifugal force acting on the accelerometer of a uniformly rotating IMU does not affect the dot product.  First, the gyroscope must be calibrated using the proposed method in this paper to obtain the scale factor. Subsequently, the data generated by the rotating accelerometer will be used to calculate the dot product with the gyroscope data. The results of both before and after calibration are shown in Figure \ref{rotatingplot}. It is evident that following the the proposed method in this paper, the dot product becomes consistent. So it can be concluded that the data generated by the accelerometer under uniform rotation has no effect on the dot product. In addition, the calibrated gyroscope is used to calculate the dot product with the data generated by the rotating accelerometer and the data generated by the stationary accelerometer. The resulting distribution diagram is shown in Figure \ref{violinplot}. From Figure \ref{violinplot} it can be observed that the mean and distribution of the dot products obtained in the two states are nearly identical. Thus, it can be concluded that the centrifugal force during uniform rotation does not significantly affect the dot product in the calibration. And the relationship between rotation speed, centrifugal force and gravitational force can be described as follows: 
\begin{equation} \label{eq_8}
	\tilde L=\boldsymbol{\omega(g+f_c)}
\end{equation}
In the equation (\ref{eq_8}), $\tilde L$ represents the dot product of rotation speed vector and the sum of the centrifugal force and the gravity vector.
The Equation (\ref{eq_8}) can be simplified to 
\begin{equation}
	\tilde L=\boldsymbol{\omega g+\omega f_c}
\end{equation}

\begin{figure}
	\centering
	    \includegraphics[width=0.5 \textwidth, height=0.26 \textheight]{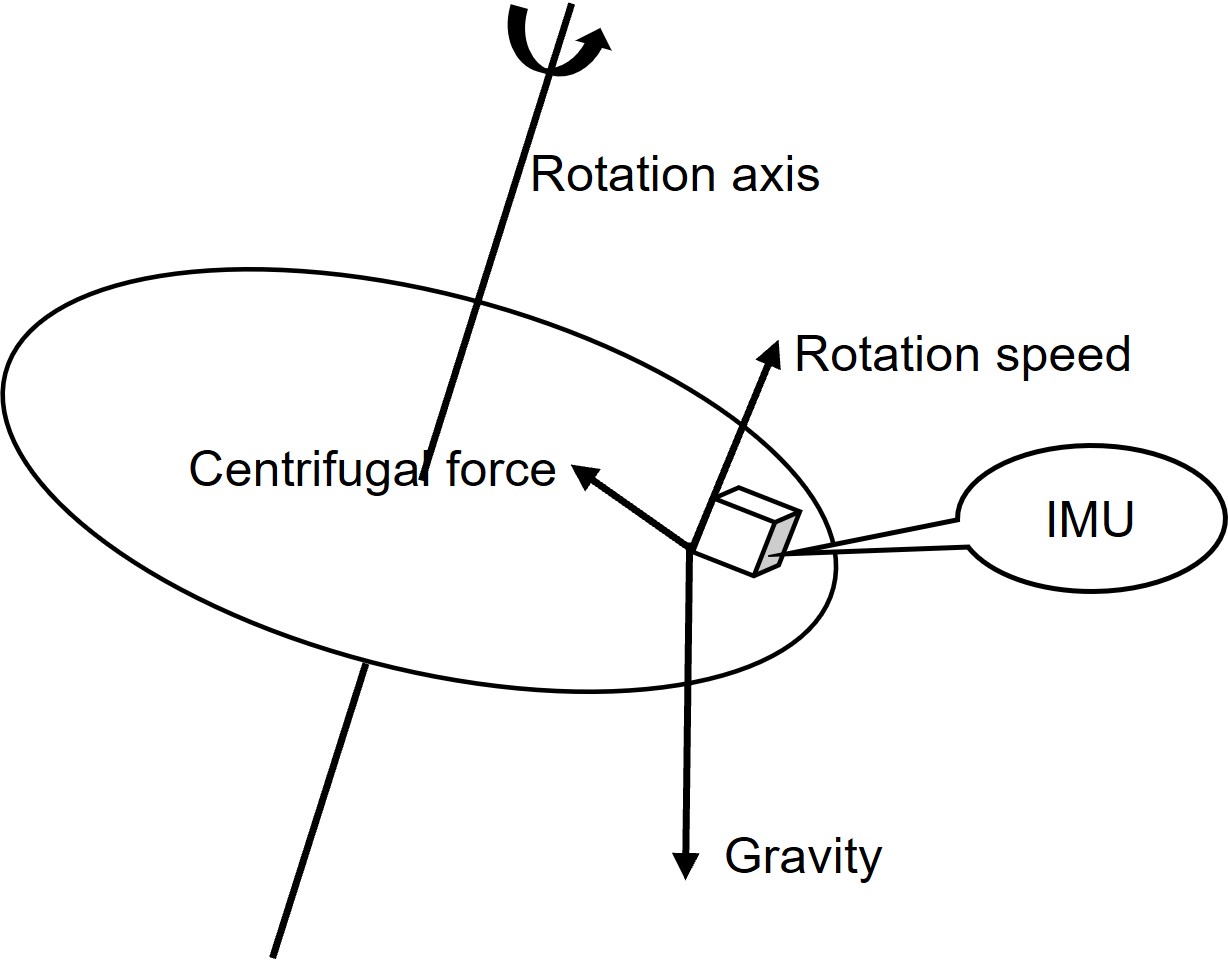}
	    \caption{The rotation speed vector, centrifugal force vector and gravity vector at constant rotation speed}
	    \label{Descripution}
\end{figure}

\begin{figure}
	\centering
	\includegraphics[width=0.5 \textwidth, height=0.25 \textheight]{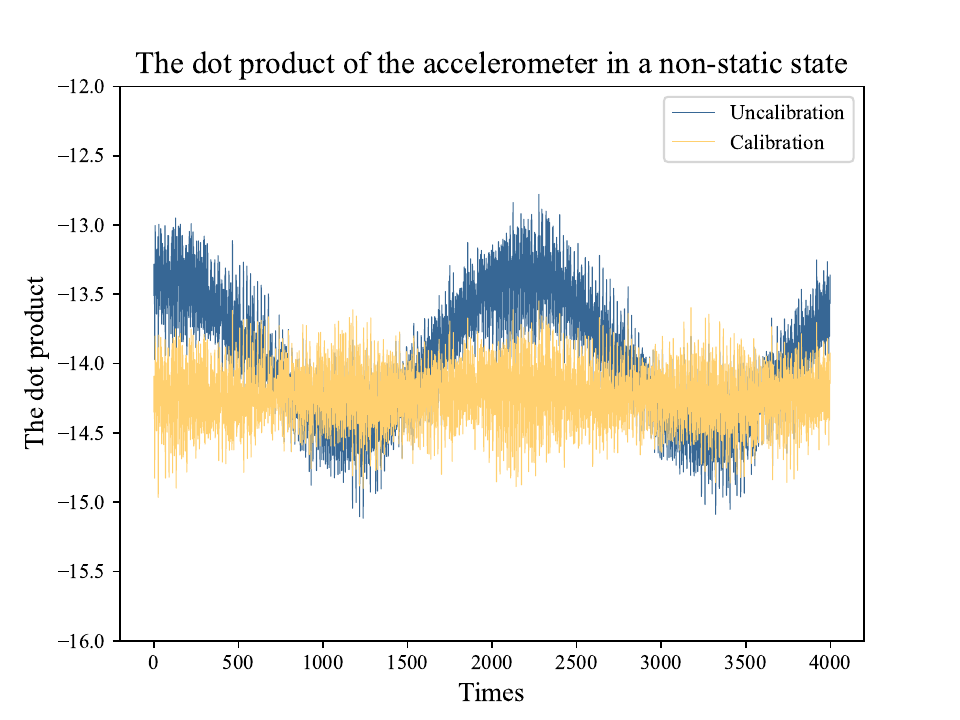}
	\caption{The dot product between accelerometer data in motion and gyroscope data}
	\label{rotatingplot}
\end{figure}

\begin{figure}
	\centering
	\includegraphics[width=0.5 \textwidth, height=0.25 \textheight]{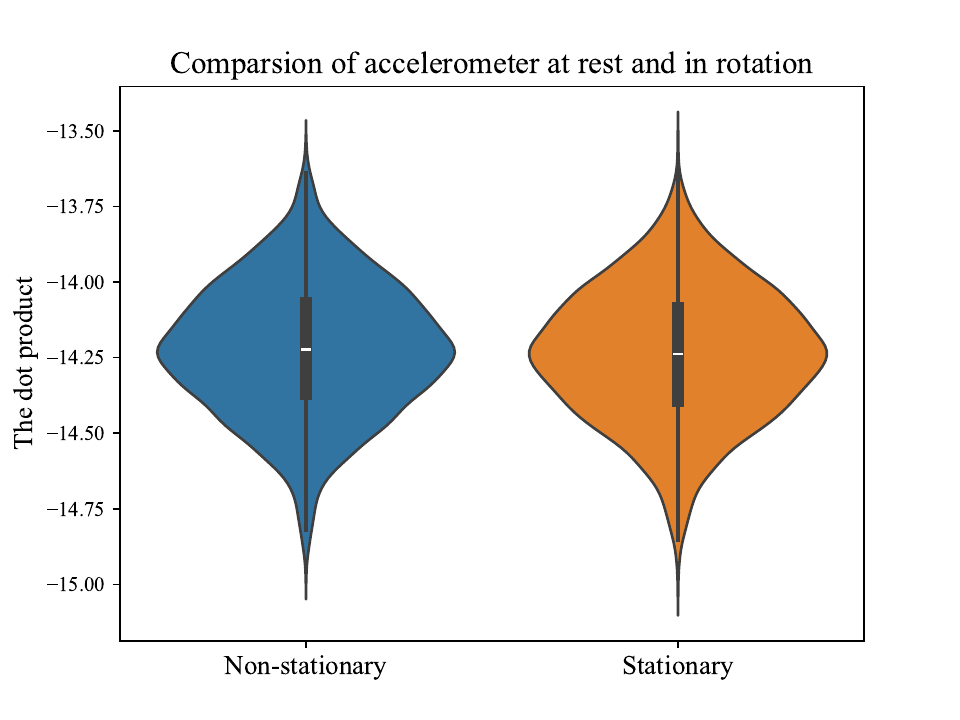}
	\caption{The distribution of the dot product in stationary and rotating accelerometer scenarios: (a) The blue violin plot delineates the distribution of dot products derived from data collected by the accelerometer and gyroscope during rotational motion. (b) The orange violin plot depicts the distribution of dot products obtained from a stationary accelerometer while the gyroscope is in rotation.}
	\label{violinplot}
\end{figure}

The relationship between the rotational vectors and acceleration vectors during calibration experiments is illustrated in Figure \ref{Descripution}. It is evident that the dot product of \( \boldsymbol{\omega} \) and \( \boldsymbol{f_c} \) is zero due to their orthogonality. Consequently, we can conclude that when the IMU rotates at a constant speed, the centrifugal force generated by this rotation does not influence the dot product. Therefore, we find that \( \tilde{L} = L \)  (see Equation (\ref{eq_2})), indicating that the dot product remains constant even when the accelerometer is in motion. In static conditions, the stochastic noise in accelerometer measurements can be mitigated by averaging the data over a period of time. However, reducing measurement noise during rotation presents a greater challenge. In future work, we will explore in detail how to effectively utilize accelerometer rotation data for calibration purposes.

 \section{Discussion}

The simulation study demonstrated the significant impact of accelerometer errors on calibration accuracy, as illustrated in Figure \ref{boxplotaccerror} and Table \ref{thespecificsimualtionresult}. Specifically, the presence of accelerometer scale factor errors was found to substantially affect the calibration outcomes. Therefore, it is recommended to first calibrate the accelerometer using Lin Ye's method \cite{ye2017efficient} to enhance the accuracy of gyroscope calibration. Additionally, the simulation investigated the effect of varying noise levels on calibration results. As depicted in Figure \ref{scalefactordistribution} and Table \ref{theinformationofscalefactor}, the influence of noise on the calibration results was minimal. Despite variances ranging from 0.01 to 10, the distribution of the scale factor remained consistent with a normal distribution, indicating the robustness of the method in the presence of noise. Furthermore, real gyroscope data, as shown in Figure \ref{scalefactordistribution} and Table \ref{distribution of actual gyro}, also exhibited a normal distribution, further validating the reliability of the method.

The experimental results corroborated the findings from the simulations, as illustrated in Figure \ref{scalefactordistribution} and Figure \ref{theboxplotrange}. These experiments were meticulously designed to facilitate rapid and accurate data collection and calibration, thereby confirming the practicality of the proposed method. The experimental setup also highlighted a significant advantage of this approach: the ability to mitigate installation errors. Since the calibration requires only that the gyroscope mass is present on each axis and the axis of rotation remains fixed, the method is both straightforward and highly effective for rapid calibration. This characteristic renders it particularly suitable for practical applications where precise and costly turntables may not be accessible.

Moreover, the dot product of the accelerometer and gyroscope data during uniform rotation was calculated, demonstrating that the centrifugal forces generated during constant-speed rotation of the accelerometer did not adversely affect calibration accuracy, as shown in Figure \ref{rotatingplot} and Figure \ref{violinplot}. This further underscores the robustness of the method in practical scenarios. The simplicity of the setup, which requires only a cost-effective servo motor and eliminates the need for secondary installation or consideration of alignment errors, ensures efficient implementation in real-world environments, particularly in resource-limited settings.

In conclusion, the simulations and experimental validations provided compelling evidence of the effectiveness of the proposed gyroscope calibration method. Its ability to mitigate the effects of noise and accelerometer errors while maintaining high accuracy establishes it as a practical solution for addressing gyroscope scale factor drift.

\section{Conclusion} This paper presents a rapid and effective methodology for calibrating gyroscopes utilizing a straightforward setup that requires only measurements from a stationary accelerometer and a rotating gyroscope. By exploiting the invariant nature of the dot product between acceleration and rotation vectors, the proposed method achieves high accuracy without necessitating expensive equipment and complex installation procedures. Both simulation and experimental validations substantiate the robustness of this approach, particularly in minimizing the effects of noise and errors associated with the accelerometer. This methodology offers an efficient and cost-effective solution for mitigating gyroscope scale factor drift in practical applications.

\section*{References}

\subsection{References}
\bibliographystyle{IEEETran}
\bibliography{IEEEabrv, main}


\end{document}